\documentclass[
twocolumn,
]{ceurart}

\usepackage{graphicx}
\usepackage{blindtext}
\usepackage{hyperref}
\usepackage{url}
\usepackage{natbib}
\usepackage{array}
\usepackage{float}

\usepackage[utf8]{inputenc}
\usepackage[english]{babel}

\usepackage{bbding}
\usepackage{tikz}

\def\checkmark{\tikz\fill[scale=0.4](0,.35) -- (.25,0) -- (1,.7) -- (.25,.15) -- cycle;} 
\DeclareOldFontCommand{\bf}{\normalfont\bfseries}{\mathbf}

\begin{document}

\copyrightyear{2020}
\copyrightclause{Copyright for this paper by its authors.
  Use permitted under Creative Commons License Attribution 4.0
  International (CC BY 4.0).}

% \conference{In:   A.  Hoppe,  R.  Yu,  Y.  Kammerer,  L.  Salmer\'{o}n:   Proceedings of the 1st International Workshop on Investigating Learning During Web Search co-located with CIKM, Galway, Ireland,19-10-2020}

\conference{Proceedings of the CIKM 2020 Workshops,  October 19--20, Galway, Ireland}

\title{Classification of Important Segments in Educational Videos using Multimodal Features}

\author[1]{Junaid Ahmed Ghauri}[%
orcid=0000-0001-9248-5444,
]
\ead{junaid.ghauri@tib.eu}
\address[1]{TIB -- Leibniz Information Centre for Science and Technology, Hannover, Germany}
\address[2]{L3S Research Center, Leibniz University Hannover, Germany}

\author[1]{Sherzod Hakimov}
[orcid=0000-0002-7421-6213]

\ead{sherzod.hakimov@tib.eu}

\author[1,2]{Ralph Ewerth}
[orcid=0000-0003-0918-6297]
\ead{ralph.ewerth@tib.eu}

\begin{abstract}
 Videos are a commonly-used type of content in learning during Web search. Many e-learning platforms provide quality content, but sometimes educational videos are long and cover many topics. Humans are good in extracting important sections from videos, but it remains a significant challenge for computers. In this paper, we address the problem of assigning importance scores to video segments, that is how much information they contain with respect to the overall topic of an educational video. We present an annotation tool and a new dataset of annotated educational videos collected from popular online learning platforms. Moreover, we propose a multimodal neural architecture that utilizes state-of-the-art audio, visual and textual features. Our experiments investigate the impact of visual and temporal information, as well as the combination of multimodal features on importance prediction.
\end{abstract}


\maketitle

%%%%%%%%%%%%%%%%%%%%%%%%%%%%%%%%%%%%%%%

\section{Introduction}

In the era of e-learning, videos are one of the most important medium to convey information for learners, being also intensively used during informal learning on the Web \citep{PardiHHK2020Role,hoppe2018current}. Many academic institutions started to host their educational content with recordings while various platforms like Massive Open Online Courses (MOOC) have emerged where a large part of the available educational content consists of videos. Such educational videos on MOOC platforms are also exploited in search as learning scenarios, their potential advantages compared with informal Web search have been investigated by \citet{MoraesPH2018Contrasting}. 
%Even though
Although many platforms pay a lot of attention to the quality of the video content, the length of videos is not always considered as a major factor. Many academic institutions provide content where the whole lecture is recorded without any breaks. Such lengthy content can be difficult for learners to follow in distant learning. As mentioned by \citet{Guo2014} shorter videos are more engaging in contrast to pre-recorded classroom lectures split into smaller pieces for MOOC. Moreover, pre-planned educational videos, talking head, illustrations using hand drawings on board or table, and speech tempo are other key factors for engagement in a video lecture as described by \citet{Zolotykhin2019/07}. 

%\begin{figure}[ht]
\begin{figure}[pos=t]
\begin{center}
\includegraphics[height=4cm]{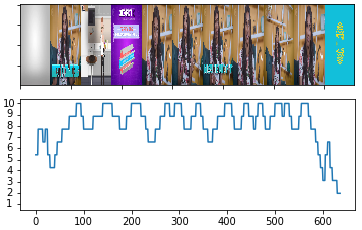}
\caption{Sample video with annotations of importance scores for each segment}
\label{fig:example}
\end{center}
\end{figure}

In this paper, we introduce computational models that predict the importance of segments in (lengthy) videos. Our model architectures  incorporate visual, audio, and text (transcription of audio) information to predict importance scores for each segment of an educational video. A sample video and its importance scores are shown in Figure~\ref{fig:example}. A value between 1 and 10 is assigned to each segment indicating the score of a specific segment whether it refers to an important information regarding the overall topic of a video. We refer to it as the \textit{importance score} of video segments in educational domain, similar to the annotations provided by TVSum dataset \citep{song2015tvsum} on various Web videos. We have developed an annotation tool that allows annotators to assign importance scores to video segments and created a new dataset for this task (see Section~\ref{sec:dataset}).
The contributions of this paper are summarized as follows:
\begin{itemize}
  \item Video annotation tool and an annotated dataset
  \item Analysis of influence of multimodal features and parameters (history window) for educational video summarization
  \item Multimodal neural architectures for the prediction of importance scores for video segments
  \item The source code for defined the deep learning models, the annotation tool and the newly created dataset are shared publicly\footnote{\url{https://github.com/VideoAnalysis/EDUVSUM}} with the research community. 
\end{itemize}

The remaining sections of the paper are organized as follows. Section 2 presents an overview of related work in video-based e-learning and computational architectures covering multiple modalities in educational domain. In Section 3, we provide detailed description of model architectures. Section 4 presents the described annotation tool and the created dataset. Section 5 covers the experimental results and discussions on the findings of the paper and Section 6 concludes the papers.

%%%%%%%%%%%%%%%%%%%%%%%%%%%%%%%%%%

\section{Related work}

Various studies have been conducted that address the quality of online education, create personalized recommendations for learners, or focus on highlighting the most important parts in lecture videos. Student interaction with lecture videos offers new opportunities to understand the performance of students or for the analysis of their learning progress. Recently, \citet{Mubarak20} proposed an architecture that uses features from e-learning platforms such as watch time, plays, pauses, forward and backward to train deep learning models for predictive learning analytics. 
% Similarly,
In a similar way, \citet{DBLP:journals/ijet/ShukorA19} used watch time, clicks, completed number of assignments for the same purpose. Another method by \citet{DBLP:conf/TangLWSCL20} is a concept-map based approach that analyzes the transcripts of videos collected from YouTube and visual recommendations to improve learning path and provide personalized content. In order to improve student performance and enhance the learning paradigm, high-tech devices are recommended for the classroom setting and content presentation. For instance, instructors or presenters can highlight important sections which can be saved along with the video data and later be used by students when they are going through the video lectures.

Research in the field of video summarization addresses a similar problem, where important and relevant content from videos is classified to generate summaries (for instance,  \cite{DBLP:conf/eccv/ZhangCSG16,DBLP:journals/tlt/YangM14} and \cite{DBLP:conf/mm/Wang000FT19}). All of these methods are based on TVSum~\citep{song2015tvsum} and SumMe~\citep{DBLP:conf/eccv/GygliGRG14} datasets that consist of Web videos. The nature of these datasets is very different to videos from the educational domain. These datasets can be a good source of visual features but spoken words or textual content are relatively rare or not present at all. Inspired from video summarization work, \citet{DBLP:conf/icdar/DavilaZ17} presented a method to detect written content in videos, e.g. on whiteboards. This research focuses on a sub-task which only takes into account the lectures in which the written content is available, and also addresses only the topic of mathematics. \citet{DBLP:conf/icdar/XuDSG19} focused on another kind of technique where speaker pose information can help in action classification like writing, explaining, or erasing. Here, the most important segments are \textit{explaining}, which could be an indication of an important segment in educational videos.

Another important aspect of e-learning is student engagement for different types of online resources. \citet{Guo2014} analyzed various aspects for MOOC videos and provided a number of related recommendations. \citet{ShiOHHE2019} analyzed the correlation of features and lecture quality by considering visual features from slides, linguistic elements and audio features like energy, frequency, pitch, etc. to highlight important and emphasized statements in a lecture video. As suggested by \citet{YukiIchimura19}, one of the best practices in MOOCs is to offer information on which parts of a lecture video are difficult or need more attention, which could potentially lead to a more flexible and personalized learning experience. In order to perform such tasks by machines, they need to incorporate multimodal information from educational content. To deal with multimodal data is not easy and this is also true for multimodal learning, as explained by \citet{Wang2019WhatMT}. If user interaction data are available for videos along with visual, textual information, then the task can be solved by multimodal deep learning models. 
% i put it in the next section, it is actually no related work
%To join different modalities we adapt and extend ideas from \cite{DBLP:journals/kbs/MajumderHGCP18} where fusion is applied to three kinds of modalities available in videos: visual, audio, and text.

%%%%%%%%%%%%%%%%%%%%%%%%%%%%%%%%%%%%%%%%

\section{Multimodal Architecture}

\begin{figure*}[]
\begin{center}
\includegraphics[width=1.0\textwidth]{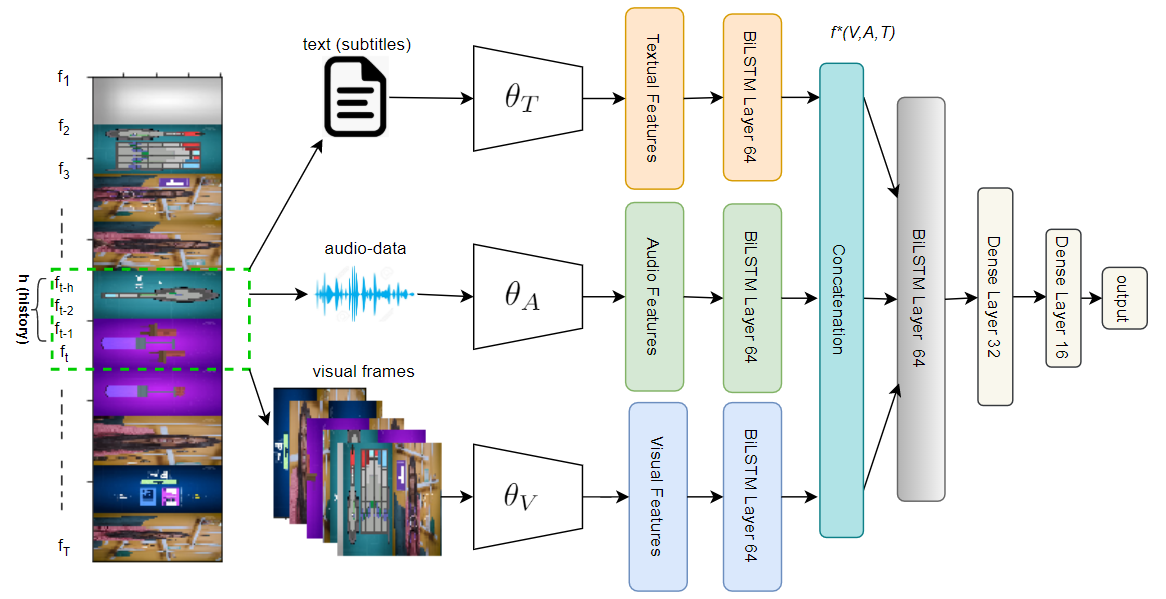}
\caption{Multimodal architecture for classification of important segments in educational videos}
\label{fig:architecture}
\end{center}
\end{figure*}

In this section, we describe the proposed model architecture that predicts importance scores for each video segment by fusing audio, visual and textual features. Each video contains audio, visual and textual (subtitles) content in the three different modalities. To join different modalities we adapt and extend ideas from \citet{DBLP:journals/kbs/MajumderHGCP18}, who apply fusion to three kinds of modalities available in videos: visual, audio, and text.
The overall architecture is depicted in Fig.~\ref{fig:architecture}.
In order to deal with the temporal aspect of videos, we use Bidirectional Long Short-term Memory (BiLSTM) layers to incorporate information from each modality \citep{Mubarak20,DBLP:conf/mm/Wang000FT19,DBLP:conf/eccv/ZhangGS18}. We use state-of-the-art pre-trained models to encode each modality in order to extract features. After the extraction of feature embeddings for each modality, they are fed into separate \emph{BiLSTM} layers. The outputs of these layers are then concatenated in a time-oriented way
% , where the concatenation axis represents temporal distribution 
and then fed into another \emph{BiLSTM} layer, which has 64 units. The output is fed into two dense layers with size of 32 and 16, respectively. Lastly, the output from the last dense layer is fed into a softmax layer that outputs a 10-dimensional vector indicating the importance score of a given input video frame belonging to a certain segment. In addition to the current frame, the model also includes history information that consists of $n$ previous frames according to the setting of history window size parameter. Our experimental results show different configurations and corresponding results, where we evaluate different history windows sizes. Next, we describe the feature embeddings for each modality and the corresponding models to extract them.

\textbf{Textual Features}: The textual content is based on subtitles provided for each video. The text features are extracted by encoding words in subtitles using BERT (Bidirectional Encoder Representations from Transformers) ~\citep{DBLP:conf/naacl/DevlinCLT19} embeddings. \emph{BERT} is a pre-trained transformer (denoted as $\theta_{T}$) that takes the sentence context into account in order to assign a dense vector representation to each word in a sentence. The textual features are 768-dimensional vectors that are extracted by encoding subtitles of videos. Later, these features are passed to a layer with 64 \emph{BiLSTM} cells.

\textbf{Audio Features}: The audio content is utilized by means of various features that represent the zero crossing rate, energy, entropy or energy, spectral features (centroid, spread, flux, roll-off) and others. In total, there are 34$\times$ $n_a$ features, where $n_a$ depends on the window size and step size which are 0.05 and 0.025 \% of the audio track length in a video. The combination of the rate of change of all these features yields a total number of 68 features. We use \emph{pyAudioAnalysis}~\citep{giannakopoulos2015pyaudioanalysis} toolkit (denoted as $\theta_{A}$) to extract these features. These features are fed into a layer with 64 BiLSTM units. We keep the same number of units in the BiLSTM layer of all modalities.

\textbf{Visual Features}: We explored different visual models like Xception~\citep{DBLP:conf/cvpr/Chollet17}, ResNet-50~\citep{DBLP:conf/cvpr/HeZRS16}, VGG-16~\citep{DBLP:journals/corr/SimonyanZ14a} and Inception-v3~\citep{DBLP:conf/cvpr/SzegedyVISW16} pre-trained on ImageNet dataset. Visual content of the videos is encoded using one of the visual descriptors mentioned above, denoted as $\theta_{V}$. Our ablation study in Section~\ref{sec:experiments} provides further details on the importance of choice of visual descriptors. Once the features are extracted, they are fed into a \emph{BiLSTM} layer with a size of 64.

Consider a video input of $T$ sampled frames, i.e., %$V=\{f_1,f_2,\ldots,f_t,\ldots,f_T\}$
$V=(f_t)\textsubscript{t=1,\ldots,T}$, $f_t$ is the visual frame at point in time t. 
%with index $t \in {1,2,\dots,T}$. 
The variable $T$ depends on the number of selected frames per second in a video. The original frame rate is 30 per second (fps) for a video. The input video is split into uniform segments of 5 seconds from which we select 3 frames per second as a sampling rate. The input of the model are the current frame ($f_t$) at time step \emph{t} and the preceding frames ($f_{t-1}$, $f_{t-2}$, \dots, $f_{t-h}$) according to the selected history window size \emph{h}. The features from a modality are extracted as defined above and passed to the respective layers. The model outputs an importance score for the given input frame ($f_t$).

%%%%%%%%%%%%%%%%%%%%%%%%%%%%%%%%%%%%%%%%

\section{Dataset and Annotation Tool}\label{sec:dataset}

\begin{figure}[pos=t]
\begin{center}
\includegraphics[height=6.2cm]{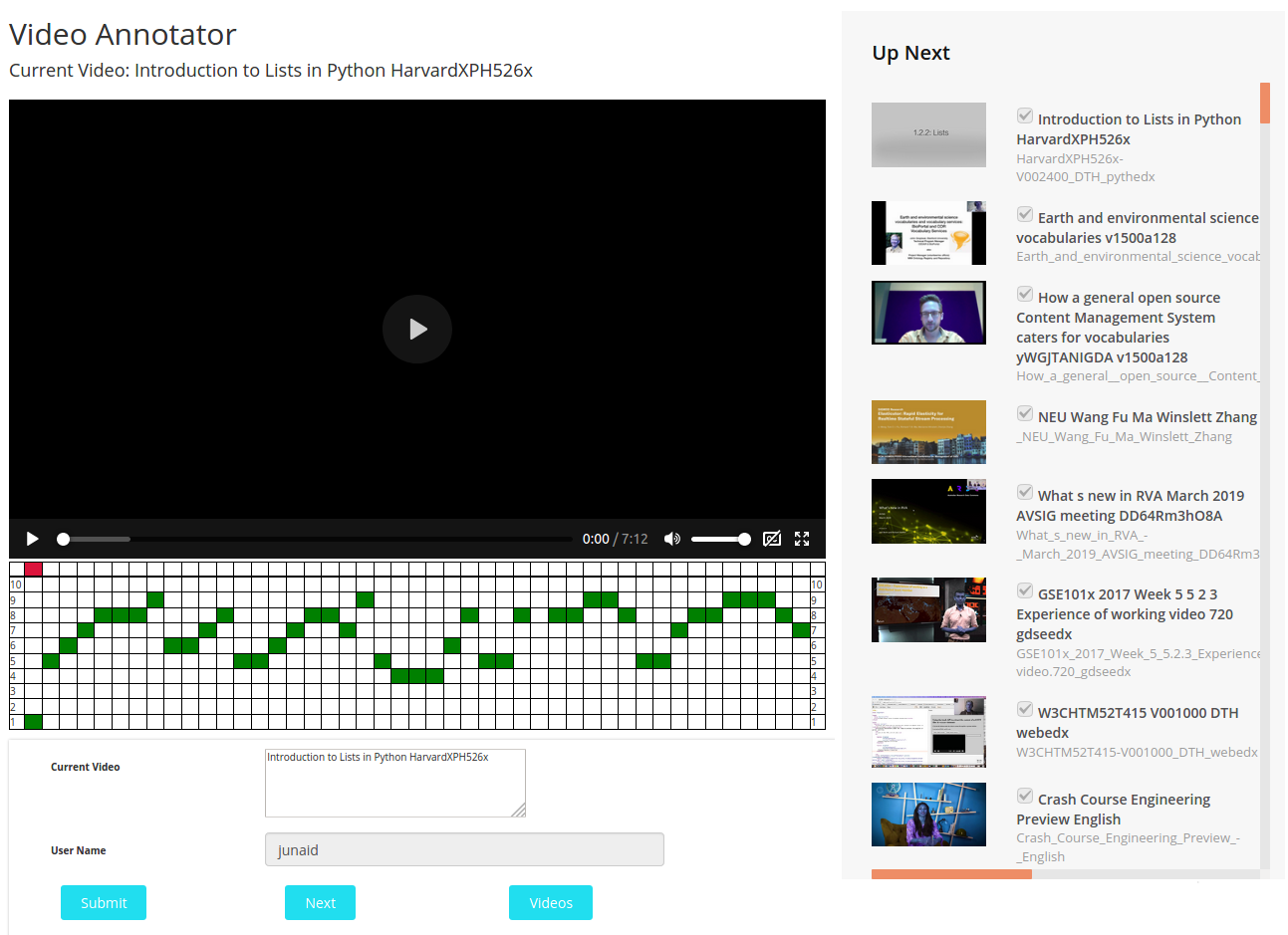}
\caption{Screenshot of the Web-based annotation tool for labeling video segments}
\label{fig:annotation_tool}
\end{center}
\end{figure}

We present a Web-based tool to annotate video data for various tasks. Each annotator is required to provide a value between 1 and 10 for every 5 second segment of a video. A sample screenshot of the annotation tool is shown in Figure~\ref{fig:annotation_tool}. The higher values indicate the higher importance of that specific segment in terms of information it includes related to a topic of a video.

We present a new dataset called EDUVSUM (Educational Video Summarization) to train video summarization methods for the educational domain. We have collected educational videos with subtitles from three popular e-learning platforms: Edx, YouTube, and TIB AV-Portal\footnote{\url{https://av.tib.eu/}} that cover the following topics with their corresponding number of videos: computer science and software engineering (18), python and Web programming (18), machine learning and computer vision (18), crash course on history of science and engineering (23), and Internet of things (IoT) (21). In total, the current version of the dataset contains 98 videos with ground truth values annotated by the main author who has an academic background in computer science. %In the current version of the dataset, we did not provide any tutorials before the annotation process.
In the future, we plan to provide annotation instructions and guidance via tutorials on how to use the software for human annotators.

\begin{figure*}[pos=t]
\begin{center}
\includegraphics[width=1.0\textwidth,height=5cm]{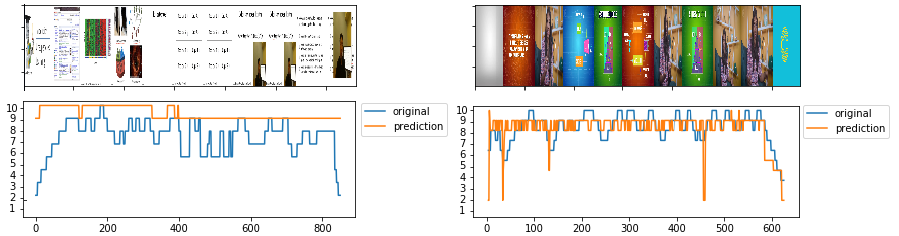}
\caption{Predictions of VGG-16 model for two videos. Left: model prediction with low accuracy (18\%), Right: model prediction with high accuracy (34\%)}
\label{fig:exampleGoodBad}
\end{center}
\end{figure*}

%%%%%%%%%%%%%%%%%%%%%%%%%%%%%%%%%%%%%%%%%%%

\section{Experimental Results}\label{sec:experiments}

In this section, we describe the experimental configurations and the obtained results. We use our newly created dataset consisting of 98 videos for the experimental evaluation of model architectures. The dataset is randomly shuffled before dividing it into disjoint train and test splits using 84.7$\%$ (83 videos) and 15.3$\%$ (15 videos), respectively. The videos are equally distributed among the topics of the dataset. The dataset splits and frame sampling strategy are compliant with previous work in the field of video summarization (\citet{DBLP:conf/eccv/ZhangCSG16}, \citet{DBLP:conf/eccv/GygliGRG14} and \citet{song2015tvsum}).

%  \textcolor{red}{We keep the number of videos almost the same from each topic mentioned early. This refrains to have biases due to any videos from the same topic. We randomly shuffle data before dividing it into a disjoint train and test set. We took this training, testing, and frame sampling strategy from pioneer work done in the field of video summarization by \citet{DBLP:conf/eccv/ZhangCSG16}, \citet{DBLP:conf/eccv/GygliGRG14} and \citet{song2015tvsum}.} Although whole dataset has more videos but we consider those in this experiment where all three modalities are available mentioned in early section. 

We evaluated different configurations of model architectures as classification and regression tasks. The experimental configurations include varying visual feature extractors, history window sizes, audio features, and textual features. In our experiments, we sampled 3 frames per second in order to not include too much redundant information where variation the between consecutive frames is low. This sampling rate corresponds to 10\% of the original frame rate of the video which has 30 frames per second. Additionally, we analyzed the effects of multimodal information by including or excluding one of the modalities. The results are given in Table~\ref{tab:table1}. All models are trained for 50 epochs over the training split of the dataset using \emph{Adam} optimizer. To avoid over-fitting we applied dropout with \emph{0.2} on \emph{BiLSTM} layers. Due to many configurations of experimental variables, we listed the best performing four models for each visual descriptor along with the respective history window sizes and input features from specific modalities or all.

Each trained model outputs an importance score for every frame in a video. We computed Top-1, Top-2 and Top-3 accuracy on the predicted importance scores of each frame by treating it as a classification task. The best performing model for Top-1 accuracy is \emph{VGG-16} with a history window size of $2$ achieving an accuracy of $26.3$, where only visual and textual features are used for training. The model with Top-2 accuracy is \emph{ResNet-50} with the history window of $3$ that is trained on visual, audio, textual features and it achieves an accuracy of $47.3$. The best performing Top-3 model is again \emph{VGG-16} with a history window $3$, visual and audio features, and it achieves an accuracy of $67.9$.

In addition, we compute the Mean Absolute Error (MAE) values for each trained model by treating the problem as a regression task. Each model listed in Table~\ref{tab:table1} includes an average MAE value based on either each frame ($avg_{fra}$) or segment ($avg_{seg}$). We performed the following post-processing in order to compare the values against ground truth where every segment (5 second window) of a video contains an importance scores between 1 or 10. As explained above, trained models output an importance scores for each frame in a video. For the calculation of $avg_{fra}$, every frame that belongs to the same segment is assigned the same value in the ground truth videos. For calculation of $avg_{seg}$, predicted importance scores of each frame belonging to the same segment are averaged. This average value is then assigned as a predicted value to a segment. The $avg_{seg}$ is an average MAE between predicted importance score of a segment and ground truth. Based on the presented results in Table~\ref{tab:table1}, the model that uses \emph{VGG-16} for visual features together with audio features and history window of $3$ performs with the least error for both frame and segment-based calculation of the average MAE.

%%% table_final

\begin{table}
  \caption{Average accuracy and Mean Absolute Error (MAE) values for different visual descriptors and history window (h) sizes. Modalities: Visual (V), Audio (A), Textual (T). $avg_{fra}$ stands for average MAE value based on all frames in a video, $avg_{seg}$ stands for average MAE for each segment in a video.}
  
  \label{tab:table1}
  
  \resizebox{0.5\textwidth}{!}{%
  
  \begin{tabular}{|*{10}{c|}}
    % \toprule
\hline
% Visual Features & h & Accuracy  & MAE  & V & A & T \\ %\cline{3-7}
% & & Top-1  & Top-2 & Top-3 & avg_{fra} & avg_{seg} &&&\\ %\hline

\multirow{2}{*}{Visual Features} & h & \multicolumn{2}{|c}{Accuracy \%} & \multicolumn{3}{|c|}{MAE}  & \multirow{2}{*}{V} & \multirow{2}{*}{A} & \multirow{2}{*}{T} \\ \cline{2-7}
& & Top-1  & Top-2 & Top-3 & $avg_{fra}$ & $avg_{seg}$ &&&\\ \hline

\multirow{6}{*}{Inception-v3} & 3 & 22.34  & 32.01  & 55.94 & 1.93 & 1.84 & \checkmark & \checkmark  & \checkmark\\ \cline{2-10}& 2 & 22.34  & 30.98  & 55.94 & 1.93 & 1.84 & \checkmark & \checkmark  & \checkmark\\ \cline{2-10}& 3 & 22.34  & 30.98  & 55.94 & 1.93 & 1.84 & \checkmark & \checkmark  & $\times$\\ \cline{2-10}& 2 & 22.34  & 47.3  & 55.94 & 1.93 & 1.84 & \checkmark & \checkmark  & $\times$\\ \cline{2-10}& 2 & 23.95  & 43.48  & 60.2 & 1.82 & 1.74 & \checkmark & $\times$  & \checkmark\\ \cline{2-10}& 3 & 23.48  & 44.07  & 64.29 & 1.73 & 1.66 & \checkmark & $\times$  & \checkmark\\ \hline

\multirow{6}{*}{VGG-16} & 1 & 22.43  & 47.29  & 66.33 & 1.92 & 1.84 & \checkmark & \checkmark  & \checkmark\\ \cline{2-10}& 2 & 22.37  & 37.47  & 57.92 & 1.87 & 1.81 & \checkmark & \checkmark  & \checkmark\\ \cline{2-10}& 3 & 25.55  & 46.19  & \textbf{67.92} & \textbf{1.51} & \textbf{1.49} & \checkmark & \checkmark  & $\times$\\ \cline{2-10}& 2 & 22.91  & 45.08  & 58.93 & 1.83 & 1.79 & \checkmark & \checkmark  & $\times$\\ \cline{2-10}& 2 & \textbf{26.26}  & 41.92  & 63.09 & 1.6 & 1.57 & \checkmark & $\times$  & \checkmark\\ \cline{2-10}& 3 & 25.65  & 41.28  & 63.21 & 1.65 & 1.62 & \checkmark & $\times$  & \checkmark\\ \hline

\multirow{6}{*}{Xception} & 1 & 23.1  & 39.13  & 57.33 & 1.88 & 1.8 & \checkmark & \checkmark  & \checkmark\\ \cline{2-10}& 3 & 22.34  & 30.98  & 55.94 & 1.93 & 1.84 & \checkmark & \checkmark  & \checkmark\\ \cline{2-10}& 2 & 22.72  & 47.17  & 59.74 & 1.88 & 1.8 & \checkmark & \checkmark  & $\times$\\ \cline{2-10}& 1 & 22.42  & 47.2  & 67.12 & 1.86 & 1.78 & \checkmark & \checkmark  & $\times$\\ \cline{2-10}& 3 & 24.04  & 37.99  & 59.76 & 1.82 & 1.74 & \checkmark & $\times$  & \checkmark\\ \cline{2-10}& 2 & 22.65  & 44.45  & 62.39 & 1.86 & 1.78 & \checkmark & $\times$  & \checkmark\\ \hline

\multirow{6}{*}{ResNet-50} & 3 & 22.6  & \textbf{47.31}  & 67.11 & 1.9 & 1.82 & \checkmark & \checkmark  & \checkmark\\ \cline{2-10}& 2 & 22.39  & 37.03  & 57.53 & 1.92 & 1.84 & \checkmark & \checkmark  & \checkmark\\ \cline{2-10}& 3 & 24.27  & 37.66  & 59.74 & 1.76 & 1.71 & \checkmark & \checkmark  & $\times$\\ \cline{2-10}& 2 & 22.75  & 37.25  & 57.34 & 1.85 & 1.81 & \checkmark & \checkmark  & $\times$\\ \cline{2-10}& 2 & 22.69  & 31.59  & 56.66 & 1.85 & 1.8 & \checkmark & $\times$  & \checkmark\\ \cline{2-10}& 1 & 22.67  & 31.61  & 57.39 & 1.81 & 1.78 & \checkmark & $\times$  & \checkmark\\   \hline

\end{tabular}
}
\end{table}

%%%%%%%%%%%%%%%%%%%%%%%%%%%%%%%%%%%%%%%%%

\subsection{Discussion}
For a deeper analysis of errors made by the trained models, we plot ground truth labels along with predictions and select two videos with relatively low (left video) and high (right video) accuracy. These plots are shown in Figure~\ref{fig:exampleGoodBad}. The video on the left side has low accuracy (18\%) because the predicted values are far off from the ground truth. The reason could be the fact that frames in the video have less visual variation and the model predicts the same or similar values for those frames. Another reason could be that the visual features are not well suited for the educational domain, since we use pre-trained models on ImageNet dataset where the task is to recognize distinct 1000 objects. On the other hand, the video on the right side has relatively high accuracy (34\%). Even though the importance scores for frames are not exact, we can observe that the model predicts lower importance scores when ground truth values are also lower, and the same pattern is observed when importance scores are increased as well. As shown in Table~\ref{tab:table1}, the best model obtains an error of $1.49$ (MAE) on average, but it is observable that most of the important segments (regardless of the predicted values) are detected by the trained model.

%%%%%%%%%%%%%%%%%%%%%%%%%%%%%%%%%%%%%%%

\section{Conclusion}

In this paper, we have presented an approach to predict the importance of segments in educational videos by fusing multimodal information. This study presents and validates a working pipeline that consists of lecture video annotation and, based on that, a supervised (machine) learning task to predict importance scores for the content throughout the video. The results show the importance of each individual modality and limitations of each model configuration. It also highlights that it is not straight forward to exploit the full potential from heterogeneous source of features, i.e., using all modalities does not guarantee a better result. 

One further direction of research is to enhance the architecture for binary and ternary fusion where modalities are fused on different levels. As a second future direction, we will focus on the release of another version of the dataset that covers more topics and videos. Finally, we will investigate other types of visual descriptors that  better fit to the educational domain.

%%%%%%%%%%%%%%%%%%%%%%%%%%%%%%%%%%%%%%%

\begin{acknowledgments}
Part of this work is financially supported by the Leibniz Association, Germany (Leibniz Competition 2018, funding line "Collaborative Excellence", project SALIENT [K68/2017]).
\end{acknowledgments}

% The project above fits better, normally we do not thank the employer here.
\begin{comment}
\begin{acknowledgments}
  Thanks to the Technische Informationsbibliothek (TIB)
German National Library of Science and Technology for supporting this research.
  \\ \url{https://www.tib.eu/en/research-development/}.
  
\end{acknowledgments}
\end{comment}

%%
%% Define the bibliography file to be used
%\bibliography{sample-ceur}

\bibliographystyle{plain}
\bibliography{references}

\end{document}